%% file: article.tex
\documentclass{article}
\usepackage{styles/arxiv}
\usepackage{import}
\usepackage[utf8]{inputenc} 
\usepackage[T1]{fontenc}    
\usepackage{hyperref}       
\usepackage{url}            
\usepackage{booktabs}       
\usepackage{amsfonts}       
\usepackage{nicefrac}       
\usepackage{microtype}      %
\usepackage{lipsum}

\usepackage{graphicx}
\usepackage{subcaption}

\usepackage{hyperref}
\usepackage{verbatim}
\usepackage{amsmath,amsfonts,amsthm,amssymb,bm,bbm}

\usepackage{MnSymbol}
\usepackage{algorithm}
\usepackage{algpseudocode}

\makeatother
\usepackage{dirtytalk}
\usepackage{float}
\usepackage{ragged2e}

\usepackage{siunitx}
\usepackage{booktabs}
\usepackage{multirow}
\usepackage{makecell}
\usepackage{booktabs}
\usepackage{tabulary}
\usepackage{graphicx}
\usepackage{comment}
\usepackage{soul}
\usepackage{xcolor}
\usepackage{multirow}
\usepackage{nicematrix}
\usepackage{algorithm}
\usepackage{soul}
\NiceMatrixOptions{
    code-for-first-row = \color{blue} ,
    code-for-last-row = \color{blue} ,
    code-for-first-col = \color{blue} ,
    code-for-last-col = \color{blue}
}

\usepackage{threeparttable}
\newcolumntype{L}[1]{>{\raggedright\arraybackslash}p{#1}}
\newcolumntype{C}[1]{>{\centering\arraybackslash}p{#1}}
\newcolumntype{R}[1]{>{\raggedleft\arraybackslash}p{#1}}

\algnewcommand\INPUT{\item[\textbf{Input:}]}
\algnewcommand\OUTPUT{\item[\textbf{Output:}]}
\algnewcommand\PARAMETERS{\item[\textbf{Parameters:}]}
\algnewcommand\GIVEN{\item[\textbf{Given:}]}
\algnewcommand\algorithmicforeach{\textbf{for each}}
\algdef{S}[FOR]{ForEach}[1]{\algorithmicforeach\ #1\ \algorithmicdo}

\title{Exploratory Analysis of Federated Learning Methods with Differential Privacy on MIMIC-III}
\import{sections/}{authors}

\begin{document}
\maketitle
\clearpage 

\import{sections/}{abstract}
\clearpage

\import{sections/}{introduction}
\import{sections/}{materials_methods}
\import{sections/}{experimental_setup}

\clearpage
\import{sections/}{results}

\clearpage
\import{sections/}{conclusions}
\clearpage
\import{sections/}{backmatter}
\clearpage

\bibliographystyle{bibliography/bmc-mathphys}
\bibliography{article} 
\clearpage

\appendix
\import{appendix/}{parametric_study}

\end{document}

%% file: sections/authors.tex
\author{
   Aron N. Horvath
   \thanks{Chair of Biomedical Informatics, University Hospital of Zurich, Zurich, Switzerland}
   \hspace{1pt}
   \thanks{These authors contributed equally to the manuscript} \\
   University Zurich \\
   \texttt{aron.horvath@uzh.ch} \\
\And
    Matteo Berchier
    \footnotemark[1]
    \hspace{1pt}
    \footnotemark[2] \\
    University Zurich \\
    \texttt{matteo.berchier@uzh.ch} \\
\And
    Farhad Nooralahzadeh
    \footnotemark[1] \\
    University Zurich \\
    \texttt{farhad.nooralahzadeh@uzh.ch} \\
\And
    Ahmed Allam
    \footnotemark[1] \\
    University Zurich \\
    \texttt{ahmed.allam@uzh.ch} \\
\And
    Michael Krauthammer
    \footnotemark[1] \\
    University Zurich \\
    \texttt{michael.krauthammer@uzh.ch}
}

%% file: sections/abstract.tex
\begin{abstract}


\textbf{Background}: 
Federated learning methods offer the possibility of training machine learning models on privacy-sensitive data sets, which cannot be easily shared. Multiple regulations pose strict requirements on the storage and usage of healthcare data, leading to data being in silos (i.e. \emph{locked-in} at healthcare facilities). The application of federated algorithms on these datasets could accelerate disease diagnostic, drug development, as well as improve patient care. 
\\
\textbf{Methods}:
We present an extensive evaluation of the impact of different federation and differential privacy techniques when training models on the open-source MIMIC-III dataset. 
We analyze a set of parameters influencing a federated model performance, namely data distribution (homogeneous and heterogeneous), communication strategies (communication rounds vs. local training epochs), federation strategies (FedAvg vs. FedProx). Furthermore, we assess and compare two differential privacy (DP) techniques during model training: a stochastic gradient descent-based differential privacy algorithm (DP-SGD), and a sparse vector differential privacy technique (DP-SVT).
\\
\textbf{Results}:
Our experiments show that extreme data distributions across sites (imbalance either in the number of patients or the positive label ratios between sites) lead to a deterioration of model performance when trained using the FedAvg strategy. This issue is resolved when using FedProx with the use of appropriate hyperparameter tuning.
Furthermore, the results show that both differential privacy techniques can reach model performances similar to those of models trained without DP, however at the expense of a large quantifiable privacy leakage.
\\
\textbf{Conclusions}:
We evaluate empirically the benefits of two federation strategies and propose optimal strategies for the choice of parameters when using differential privacy techniques.

\end{abstract}

\vspace*{0.5cm}
\keywords{
federated learning \and 
federated averaging \and
FedAvg \and 
federated proximity \and
FedProx \and
differential privacy \and 
stochastic gradient descent \and
sparse vector technique \and
DP-SVT \and
DP-SGD \and 
timeseries modelling \and
MIMIC-III
}

%% file: sections/introduction.tex
\section{Introduction}
In the past decade, new regulations emerged aimed at protecting consumers' data rights (collection and usage) and avoiding the increasing number of data breaches and unrestrained data collection by large corporations. These regulations, most notably the General Data Protection Regulation (GDPR)\cite{GDPR2016}, as well as the Health Insurance Portability and Accountability Act (HIPAA)\cite{HIPAA1996} have led companies and institutions to search for new ways of leveraging their data without the need to move/share the data outside of their internal networks. 
Federated learning, which emerged from research in distributed learning systems, such as the work of McMahan et al. \cite{McMahan2016}, has been adopted as a method to avoid infringing on the regulations mentioned above.
Federated learning, allows the training of a global model by coordinating the training procedure of multiple models residing on separate sites (e.g., devices, silos), thus avoiding the need to transfer data outside of an organization's network. It was initially used to train a model for next-word prediction on smartphones' keyboards using millions of cell phones \cite{McMahan2016}. Today federated learning is also used in the context of cross-silo training, where siloed data found in hospitals, banks, health insurers, etc., can be leveraged to train machine learning models.

\subsection{Federated Learning (FL)}
Many adaptations to the original federated averaging strategy \cite{McMahan2016} have been proposed in the literature.
Konecny et al. presented a new optimization algorithm for training federated learning models, the so-called Federated Stochastic Variance Reduced Gradient (FSVRG)\cite{Konecny2016-1}. This algorithm modifies the local gradient calculation to obtain an unbiased estimation of the gradient. They tested it on a binary classification problem (i.e. predicting if a post would get at least a comment) using  a dataset of public Google+ posts. They showed that FSVRG could achieve faster convergence when compared with Federated SGD. 
Another proposed strategy is the Federated Proximity (FedProx) approach by Li et al.\cite{Li2018} that aimed at tackling the problem of heterogeneity of data across sites. They modified the loss calculation on the sites, by regularizing it with a proximity term (proximity of the local model weights to the federated model weights). They showed that FedProx helps with performance convergence and outperforms FedAvg in situations where sites have highly heterogeneous data distributions (such as non-IID data).

\subsection{Differential Privacy (DP)}
Federated learning enables training of machine learning models on distributed datasets without the need for centralization (sharing of personal data with the server). Although this protects the client's data from eavesdropping as well as potential malicious servers, private information can still be extracted/reconstructed from the shared model weights or directly from the final model through so-called inference attacks. These attacks are aimed at identifying whether a specific sample was part of the training or even completely reconstructing the sample itself \cite{Melis2018, Geiping2020, Boenisch2021} leading to severe privacy leakage.
To prevent such privacy leakage, one may use differential privacy, which artificially adds noise to the model training procedure (either directly to the samples or to the model weights) to protect privacy. Differential privacy, as introduced by Dwork et al. provides a framework for quantifying the privacy loss for any given sample \cite{Dwork2013} involved in the training of a machine learning model. 

Stochastic gradient-based differential privacy (DP-SGD) algorithms provide a way to perturb the training procedure by injecting noise during the back-propagation step  \cite{Abadi2016}. Another algorithm for DP is the sparse vector technique (SVT), which applies noise to the model parameters using selection criteria and releases only a portion of model weights (Q) to the central server \cite{LiW2019, Lyu2017}. Li et al. explored the different parameters of the SVT algorithm and their impact on model's performance for the federated brain tumor segmentation task \cite{LiW2019}.


\subsection{Federated Learning on MIMIC-III}
Benchmark datasets are widely used to evaluate novel machine learning solutions against state-of-the-art approaches. In the medical domain, one of the most commonly used dataset is the Medical Information Mart for Intensive Care III (MIMIC III) database offering 4 different tasks from binary classification to multivariate regression \cite{Johnson2016,Harutyunyan2019}. In recent years, this data set - particularly the in-hospital mortality prediction task -  was also used in studies demonstrating new federated learning frameworks, however without providing an extensive assessment of both the aggregation/federation strategies and the differential privacy approaches on models training and performance.
Sharma et al. showed that federated learning can achieve similar performance as centralized training using only two simulated sites developed with the coMind federated learning toolkit \cite{Sharma2019}.
Lee et al. deployed a custom federated framework on Amazon Web Services as a proof of concept and presented federated learning on balanced and imbalanced data distribution with three simulated sites achieving comparable accuracy, and precision to the state-of-the-art centralized training \cite{Lee2020}.
Budrionis et al. presented a PySyft-based federated learning solution deployed on a Google Compute Engine both with virtual and real workers \cite{Budrionis2021}. In their work, they assessed three different experiment scenarios: (1) simulating different amounts of data availability on a fixed number of sites; (2) assessing the influence of the number of sites; and (3) different data distribution across the sites not only considering the federated model performance but providing an additional benchmark on time-cost of communication rounds and inference.
Sadilek et al. demonstrated that federated learning with 20 simulated sites can achieve similar accuracy compared to centralized training federated learning and in contrast to the previous work it considered differential privacy as a privacy-preserving mechanism \cite{Sadilek2021}.
Lastly, Choudhury et al. also presented a differential-privacy enabled federated learning and evaluated the impact of differential privacy with different levels of guarantees on the final federated model performance \cite{Choudhury2019}.

\subsection{Goal / Contribution}
In this paper, we use federated learning to train machine learning models on the Medical Information Mart for Intensive Care III (MIMIC III) data set \cite{Johnson2016}. We focus on the benchmark task of in-hospital mortality prediction presented by Harutyunyan et al. \cite{Harutyunyan2019}. We aim to provide a comprehensive evaluation of the two most common federation strategies (FedAvg, FedProx) and two differential privacy algorithms (DP-SGD, DP-SVT) with respect to the final model performance and convergence. Furthermore, we test different data distribution setups, both homogeneous (varying number of samples per site)and heterogeneous (class imbalance of positive labels across the different sites).

%% file: sections/materials_methods.tex
\section{Materials and Methods}

\subsection{Federated Learning Strategies}
Federated learning algorithms are defined by their aggregation strategy and whether or not they influence the local training procedure. The aggregation strategy defines how the model weights from different sites are to be considered in the calculation of the federated model weights. In this work, we focus on the two most commonly used aggregation strategies, federated averaging (FedAvg) and federated proximity (FedProx). 

\subsubsection{Federated Averaging (FedAvg)}
The original federation strategy proposed by McMahan et al. \cite{McMahan2016}, called federated averaging (FedAvg) uses a weighted averaging of the model weights returned by the sites to calculate the federated model weights. The weighting coefficient is the number of samples used at each site for model training divided by the total number of samples across all sites. Eq. \ref{eq:FedAvg} shows the averaging procedure ($w$ are the model weights, $N^{(i)}$ is the number of samples at the $i$-th specific site and $K$ is the total number of sites):

\begin{equation}
\label{eq:FedAvg}
w_{federated} = \frac{1}{\sum_{i=1}^{K} N^{(i)}} \sum_{i=1}^{K} N^{(i)} w_{site}^{(i)}
\end{equation}

This federation strategy shows promising results for cases with clients having IID data sets. However, this strategy biases the final federated model towards the model trained on the client with the most samples. In their paper, McMahan et al. \cite{McMahan2016} tested the FedAvg strategy on an image classification problems (MNIST and CIFAR-10 image classification) and language model for next character prediction (dataset built from The Complete Works of William Shakespeare). The authors use a convolutional neural network and recurrent neural network (long-short term memory - LSTM\cite{Hochreiter1997}) for their model architectures. They generated high-quality trained models for both tasks using FedAvg, and showed that the training procedure works even with few rounds of communication between local sites and a central server. However, one of the main drawbacks of federated averaging is that the \emph{unconstrained} training procedure on each site leads the model weights to diverge considerably from the federated model weights (in particular for scenarios where the sites have different data label distributions such as non-IID data scenario).

\subsubsection{Federated Proximity (FedProx)}
Federated proximity (FedProx), proposed by Li et al.\cite{Li2018}, aims to address the main drawback of federated averaging mentioned above. This federation strategy constrains the training procedure on the sites to reduce the divergence of the trained model weights from the federated model weights (so-called \emph{client drift}). The authors propose the addition of a regularization term (i.e., L2-norm squared) in the loss calculation on the site. 
This regularization term can lead to a slower convergence rate of the federated model training, however, it is expected to smooth the convergence curves as updates to the federated model are smaller and more consistent across sites. Eq. \ref{eq:FedProx} shows how the local loss calculation on the site is modified to reduce client drift at $i$-th communication round and$j$-th local training epoch ($w$ are the model weights and $\mathcal{L}$ is the loss computed on the site for backpropagation):

\begin{equation}
\label{eq:FedProx}
\mathcal{L}_{site}^{(j)} = \mathcal{L}_{site}^{(j)} + \frac{1}{2} \mu \lVert w_{site}^{(j)} - w_{federated}^{(i)} \rVert ^{2}
\end{equation}

When the regularization parameter $\mu$ is set equal to 0, FedProx is equivalent to FedAvg. It's important to mention that the regularization procedure on the site can be used in combination with any aggregation strategy on the server-side.

Table \ref{table:FederationStrategies} shows the key differences between the presented federation strategies:

\begin{table}[h]
\centering
\begin{tabular}{lcc} 
     \toprule
     \textbf{Domain} & \textbf{FedAvg} & \textbf{FedProx} \\
     \toprule
     \textit{Model merging} & \makecell{Weighted averaging of\\model weights} & \makecell{Weighted averaging of\\model weights} \\
     \textit{Local training procedure} & $
 \mathcal{L}_{site}^{(j)}$ & \makecell{ $
 \mathcal{L}_{site}^{(j)} + \frac{1}{2} \mu \lVert w_{site}^{(j)} - w_{federated}^{(i)} \rVert ^{2}$}\\
     \bottomrule
\end{tabular}
\vspace{1ex}
\caption{Comparison of federation strategies.}
\label{table:FederationStrategies}
\end{table}

\subsection{Differential Privacy (DP)}
\label{sec:mm:differential_privacy}
Although federated learning can preserve the local data privacy to a certain extent, it can be vulnerable to malicious acts such as model inference attacks leading to privacy leakage \cite{Melis2018}. For example, inference attacks can reveal properties of the training samples such as sex, age or race \cite{Melis2018}, or even partially reconstruct the original images up to a certain level where someone can recognize its content \cite{Geiping2020}. One approach to ameliorate these issues is the use of privacy-preserving mechanisms, such as differential privacy (DP), to reduce the risk of privacy leakage in federated learning systems \cite{Yin2021}.

In this paper we focus on differential privacy-based algorithms. The formal concept of differential privacy (DP) was introduced by Dwork et al. \cite{Dwork2013}. Their formal definition of a differentially private algorithm is given by Eq. \ref{eq:DP_Definition}. An algorithm is $(\varepsilon, \delta)$-differentially private, if:

\begin{equation}
\label{eq:DP_Definition}
\text{Pr} \left[ \mathcal{M}(D) \in \mathcal{S} \right] \leq \exp{(\varepsilon)} \text{Pr} \left[ \mathcal{M}(D') \in \mathcal{S} \right] + \delta
\end{equation}

holds for any model output $\mathcal{S} \subseteq \text{Range}(\mathcal{M})$ and for any two data sets $D$ and $D'$ differing only in a single item. If $\delta = 0$, then the algorithm $\mathcal{M}$ is said to be $\varepsilon$-differentially private. Dwork et al. \cite{Dwork2013} provide an interpretation of the privacy guarantees that differential privacy offers to individual samples in a data set (e.g., patient), namely \textit{"... differential privacy promises that the probability of harm was not significantly increased by their [patient] choice to participate [in the data set]..."} \cite{Dwork2013}. 

Differential privacy algorithms can be broadly divided into two main categories: (1) global/central differential privacy, and (2) local differential privacy. Global differential privacy is used to avoid information leaking from one site to another through the merging procedure on the central server (malicious site attack). An example is when one site provides malicious model weights updates that can be leveraged to reconstruct another site model weights. On the other hand, local differential privacy is used to avoid information leaking from one site to the central server (malicious server attack). 


In this paper, we focus on local differential privacy approaches using (1) stochastic gradient descent-based DP (DP-SGD), and (2) the sparse vector technique DP (DP-SVT).

\subsubsection{Stochastic Gradient Descent-based Differential Privacy (DP-SGD)}
The method was proposed by Abadi et al. \cite{Abadi2016} and is widely implemented in various Python packages, such as Opacus for PyTorch \cite{Opacus} and TensorFlow Privacy for TensorFlow \cite{TensorFlowPrivacy}. DP-SGD is an attractive method for applying differential privacy to deep learning models since it closely mimics the classic stochastic gradient descent-based training of neural networks and is applicable to almost all network architectures. 
This method directly introduces differential privacy during the model training by perturbing the model gradients. To achieve this, the DP-SGD algorithm calculates per sample gradients, clips them to a predefined threshold, and then aggregates them into a batch gradient. As a final step, DP-SGD adds Gaussian noise to the batch gradient before updating the model parameters (Alg. \ref{alg:DP-SGD}).

\begin{center}
\begin{minipage}{0.9\linewidth}
\begin{algorithm}[H]
\caption{Federated Stochastic Gradient Descent Differential Privacy (DP-SGD) \cite{Abadi2016}}
\label{alg:DP-SGD}
\begin{algorithmic}[1]
\PARAMETERS Noise multiplier $\sigma$, gradient norm threshold $\gamma$
\GIVEN Number of batches $B$, $j$-th batch samples $\{x_1^{(j)},..., x_N^{(j)}\}$, learning rate $\eta_j$, loss function $\mathcal{L}(\textbf{w}, x)$
\INPUT Federated model weights $\textbf{w}_{federated}$

\State Starting model weights: $\textbf{w}_0 \gets \textbf{w}_{federated}$
\ForEach{batch $j \in 1,..., B$}
    \ForEach{batch sample $i \in 1,..., N$}
        \State Compute sample gradient $\textbf{g}_{i,j-1} \gets \nabla_{\textbf{w}_{j-1}} \mathcal{L} \left( \textbf{w}_{j-1}, x_i^{(j)} \right)$
        \State Clip sample gradient $\bar{\textbf{g}}_{i,j-1} \gets clip \left( \textbf{g}_{i,j-1}, \gamma \right)$
    \EndFor
    \State Compute batch gradient by adding noise $\tilde{\textbf{g}}_{j-1} \gets \frac{1}{N} \left( \sum_i \bar{\textbf{g}}_{i,j-1} + \mathcal{N} \left( 0, \sigma^2 \gamma^2 \textbf{I} \right) \right)$
    \State Gradient descent $\textbf{w}_{j} \gets \textbf{w}_{j-1} - \eta_j \tilde{\textbf{g}}_{j-1}$
\EndFor
\OUTPUT Differentially private model weights $\textbf{w}_{B}$

\end{algorithmic}
\end{algorithm}
\end{minipage}
\end{center}

\subsubsection{Sparse Vector Technique Differential Privacy (DP-SVT)}
We implemented the algorithm proposed by Li et al. \cite{LiW2019}, that combines selective parameter update (i.e., sharing only a subset of all model weights with the server) with the sparse vector technique (SVT) \cite{Lyu2017}. The sparse vector technique works according to the algorithm described in Alg. \ref{alg:DP-SVT}, where $Lap(x)$ is the Laplace distribution with location parameter $\mu=0$ and scale parameter $b=x$.

\begin{center}
\begin{minipage}{0.9\linewidth}
\begin{algorithm}[H]
\caption{Sparse Vector Technique (SVT)\cite{LiW2019}}
\label{alg:DP-SVT}
\begin{algorithmic}[1]
\PARAMETERS Privacy budgets for gradient query $\varepsilon_1$, threshold $\varepsilon_2$ and answer $\varepsilon_3$
\PARAMETERS Gradient clipping threshold $\gamma$, portion to release $Q$, local training epochs $N$
\INPUT Model weights updates $\Delta \textbf{w} = \textbf{w}_{local}^{final} - \textbf{w}_{federated}$

\State Normalize weight updates by epochs: $\Delta \bar{\textbf{w}} = \Delta \textbf{w} / N$
\State Sensitivity: $s \gets 2 \gamma$  \Comment{Default value}
\State Number of releasable weights: $q \gets size ( \Delta \bar{\textbf{w}} ) \cdot Q$
\State Privacy budget for threshold: $\varepsilon_2 \gets \left( 2 q s \right)^{\frac{2}{3}} \cdot \varepsilon_1$ \Comment{Default value}
\State Threshold: $\tau \gets Percentile_{Q} \left( \lvert \Delta \bar{\textbf{w}} \rvert \right)$
\State Noisy threshold: $\hat{\tau} \gets \tau + Lap \left( \frac{s}{\varepsilon_2} \right)$
\State List of releasable weight updates: $\Delta \hat{\textbf{w}} \gets \emptyset$

\While{$size (\Delta \hat{\textbf{w}}) \less q$}
    \State Choose a random weight update $\Delta \bar{w}_i$ from $\Delta \bar{\textbf{w}}$
    \State Compute the weight update query: $\Delta w_i^{query} \gets clip(\lvert \Delta \bar{w}_i \rvert, \gamma) + Lap \left( \frac{qs}{\varepsilon_1} \right)$
    \If{$\Delta w_i^{query} \geq \hat{\tau}$}
        \State Compute the weight update answer: $\Delta w_i^{answer} \gets clip \left( \Delta \bar{w}_i + Lap \left( \frac{qs}{\varepsilon_3} \right), \gamma \right)$
        \State Release the answer by appending $\Delta w_i^{answer}$ to $\Delta \hat{\textbf{w}}$
    \EndIf
\EndWhile
\State Rescale weight updates: $\Delta \textbf{w}_{release} \gets \Delta \hat{\textbf{w}} \cdot N$
\OUTPUT Release weight updates $\Delta \textbf{w}_{release}$
\end{algorithmic}
\end{algorithm}
\end{minipage}
\end{center}

The sparse vector technique for differential privacy (DP-SVT) is executed only once per communication round in a federated scenario and is thus considerably faster to compute than its DP-SGD counterpart. Furthermore, DP-SVT can be applied to model weights trained with a number of different optimization algorithms beyond stochastic gradient descent, as it makes no assumption on how the weights deltas are computed.

\subsection{Dataset - Medical Information Mart for Intensive Care III (MIMIC III)}

The MIMIC-III database contains data collected during 53,423 separate hospital admissions, covering 38,597 adult patients (above 16 years of age) at the Beth Israel Deaconess Medical Center in Boston between 2001 and 2008 \cite{Johnson2016}. The dataset is a staple of machine learning research in healthcare where Harutyunyan et al. presented a set of four benchmark tasks for timeseries modeling on the MIMIC-III data, consisting of (a) in-hospital mortality prediction, (b) decompensation prediction, (c) phenotyping and (d) length of stay prediction along with open-sourced data pre-processing pipeline for reproducibility \cite{Harutyunyan2019}. For the scope of this paper, we decided to focus on the benchmark task of (a) in-hospital mortality prediction. \\

\textbf{Benchmark Task - In-hospital Mortality Prediction}

For this benchmark, Harutyunyan et al. use a selection of 17 clinical variables extracted from the CHARTEVENTS and LABEVENTS tables of the MIMIC-III dataset \cite{Johnson2016,Harutyunyan2019}. The variables and their type are listed in Table \ref{table:MIMICIII_Features}. Given a sequence of measurements (i.e. features in Table \ref{table:MIMICIII_Features}) collected during the first 48 hours of a patient's stay in the ICU, the aim is to predict in-hospital mortality within 30 days of ICU admission.

\begin{table}[h]
\centering
\begin{threeparttable}
\begin{tabular}{llcc}
    \toprule
    \textbf{Type} & \textbf{Variable} & \textbf{Modeled as} & \textbf{\# of features} \\
    \toprule
    Standard measure & Height & Continuous & 1 \\
    Standard measure & Weight & Continuous & 1 \\
    \midrule
    Vital sign & Temperature & Continuous & 1 \\
    Vital sign & Heart rate & Continuous & 1 \\
    Vital sign & Mean blood pressure & Continuous & 1 \\
    Vital sign & Diastolic blood pressure & Continuous & 1 \\
    Vital sign & Systolic blood pressure & Continuous & 1 \\
    Vital sign & Capillary refill rate & Categorical & 2 \\
    Vital sign & Oxygen saturation & Continuous & 1 \\
    Vital sign & Respiratory rate & Continuous & 1 \\
    Vital sign & Fraction inspired oxygen & Continuous & 1 \\
    \midrule
    Laboratory test & Glucose & Continuous & 1 \\
    Laboratory test & pH & Continuous & 1 \\
    \midrule
    Score / scale & Glasgow coma scale - Eye opening & Categorical & 8\tnote{*} \\
    Score / scale & Glasgow coma scale - Verbal response & Categorical & 12\tnote{*} \\
    Score / scale & Glasgow coma scale - Motor response & Categorical & 12\tnote{*} \\
    Score / scale & Glasgow coma scale - Total & Categorical & 13 \\
    \bottomrule
\end{tabular}
\vspace{1ex}
\begin{tablenotes}
\scriptsize
\item[*] According to the official Glasgow coma scale \cite{GlasgowComaScale} there are 5 categories for eye opening, 6 categories for verbal response and 7 categories for motor response
\end{tablenotes}
\vspace{1ex}
\caption{Features used in the MIMIC-III in-hospital mortality benchmark task \cite{Harutyunyan2019}. For each feature an additional mask is calculated to represent if a features was measured (true) or imputed (false), leading to a total of 76 features.}
\label{table:MIMICIII_Features}
\end{threeparttable}
\end{table}

The pre-processing procedure, including the filtering criteria for patients to be included, is described in great detail in Harutyunyan et al. \cite{Harutyunyan2019}.

\subsection{Modelling - Neural Network Architecture}
Patients' data were represented as a set of multivariate timeseries describing their trajectories up to the time of prediction.
In a first step, for a given patient's sequence, we embedded the input features at each time point using an embedding layer of size equal to 16 neurons followed by a rectified linear activation function (ReLU) and a dropout layer with probability $p=0.3$. Then the embedded representation at each time point is fed to a bidirectional recurrent neural network with long-short term memory (RNN-LSTM)  cells \cite{Hochreiter1997} to capture the sequential/time dependency among the embedded features. A second dropout layer was used with probability $p=0.3$. Lastly, the output layer used an embedding layer (affine transformation) followed by $sigmoid$ activation function to compute the probability of the target outcome (i.e. in-hospital mortality). 
For model training, we used the  Adam optimizer \cite{Kingma2015} with an initial learning rate of 0.001 and a batch size of 64 as fixed hyperparameters.

%% file: sections/experimental_setup.tex
\section{Experimental Setup}

The simulated scenarios consist of two sites (simulating two separate hospitals) and a federation server. The data used on the two sites is based on the MIMIC-III benchmark task. We systematically varied the distribution of data between the two sites, to simulate increasingly unbalanced scenarios in two separate cases: changing (1) the number of samples per site and (2) the number of samples with positive label per site. 
Additionally, we evaluated two federated learning strategies (FedAvg and FedProx), as well as two differential privacy methods (DP-SGD and DP-SVT). 
We assessed the models performance on a held-out test set (3'236 patients) and using bootstrapping (resampling 10'000 times the test set) to compute the average area under the ROC (AUROC) and the precision-recall curves (AUPRC) metrics. 
In order to provide a baseline for performance comparisons, we trained the same neural network exclusively on the available local data in each configuration. 

\begin{figure}
    \begin{center}
        \includegraphics[width=\textwidth]{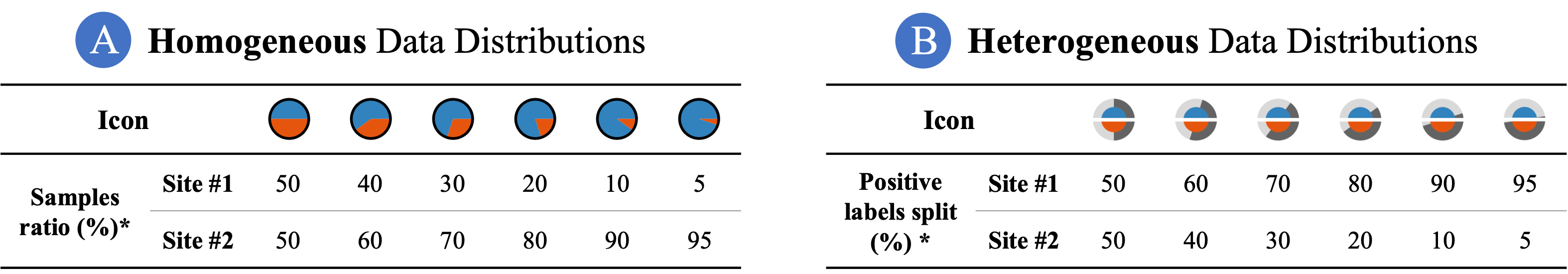}
        \vspace{5pt}
        \includegraphics[width=\textwidth]{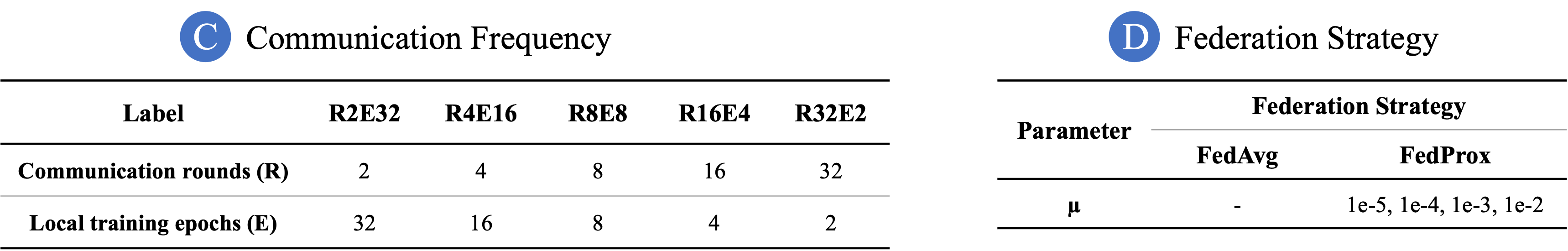}
        \vspace{5pt}
        \includegraphics[width=0.8\textwidth]{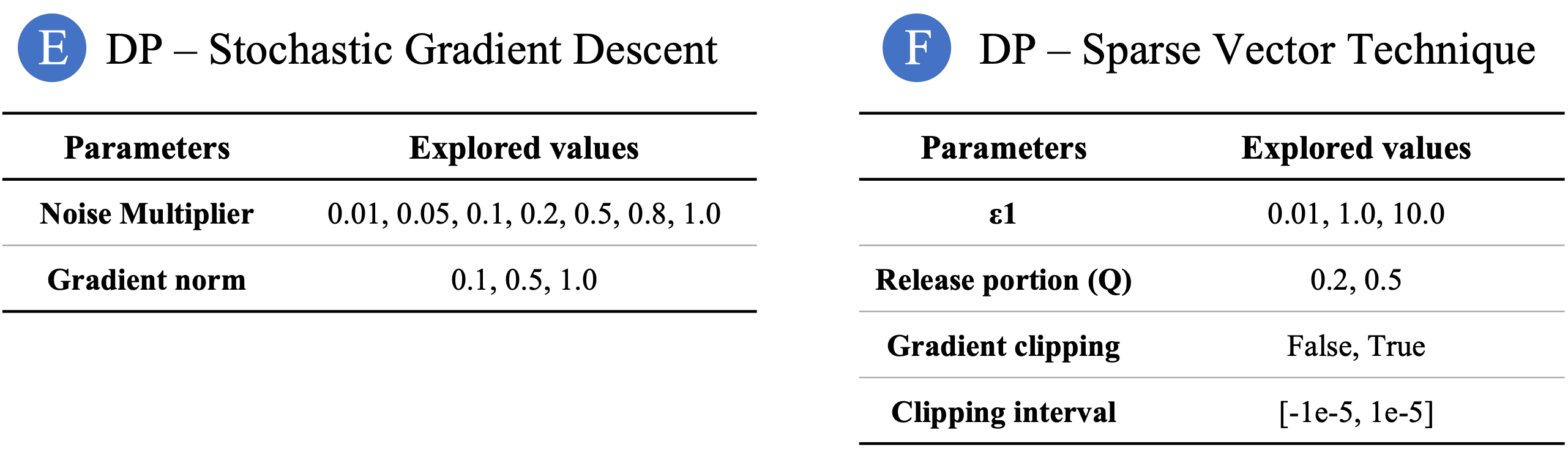}
    \end{center}
    \caption[Parametric Study Overview]{
        Overview of explored parameters.
        \\
        \textbf{Data Distributions}: 
        \textit{(A) Homogeneous}: The number of samples at each site is progressively changed from 50-50\% to 5-95\%, while the negative to positive labels ratio is kept constant. 
        \textit{(B) Heterogeneous}: The split of positive labels across the two sites is progressively increased from 50-50\% to 95-5\%, while the total number of patients per site is kept constant (50\% in Site \#1 and 50\% in Site \#2).
        \\
        \textbf{Federation Strategies}:
        \textit{(C) Communication Frequency}: The number of used communication rounds and the corresponding number of local training epochs. All scenario reach a total of 64 accumulated epochs.
        \textit{(D) Aggregation Algorithms}: federated averaging (FedAvg) and federated proximity (FedProx) are evaluated on data distributions in tables A-B. For federated proximity we explore different values of the regularization parameter $\mu$.
        \\
        \textbf{Differential Privacy}:
        \textit{(E) Stochastic Gradient Descent (DP-SGD)}: The noise perturbation is continuously increased through the noise multiplier parameter. We use different gradient norms.
        \textit{(F) Sparse Vector Technique (DP-SVT)}: The noise parameter $\epsilon_1$ is continuously increased, while the number of shared model parameters is also varied (portion to release Q going from 20 to 50\%). Gradient clipping is used for further regularization of the updated weights shared with the server. 
    }
    \label{fig:ParametricStudy_Overview}
\end{figure}

\subsection{Data Distributions}
We considered two data distribution scenarios, homogeneous and heterogeneous. In the homogeneous data distribution case, the sites only differ in the available amount of data (i.e., number of patient samples), while keeping the share of patients with a positive label (i.e., in-hospital mortality in the MIMIC-III case) constant. Figure \ref{fig:ParametricStudy_Overview}-A shows the exact data splits chosen for the parametric study and the corresponding icons used in the follow-up figures. 

In the heterogeneous data distribution case, the number of patients samples per site is kept constant, while the patient samples with a positive label are split between the two sites in increasingly unbalanced ratios, 50-50\% up to 95-5\% (Figure \ref{fig:ParametricStudy_Overview}-B). For these experiments we mainly used the AUPRC metric for evaluation.

\subsection{Communication Rounds}
To evaluate the impact of communication rounds (R) and number of local training epochs (E) on the model performance, we trained federated models on each of the data distributions scenarios (i.e. homogeneous and heterogeneous) separately with the following communication rounds/epoch configurations: 2/32, 4/16, 8/8, 16/4, 32/2 (Figure \ref{fig:ParametricStudy_Overview}-C).

\subsubsection{Aggregation Methods}
We compare the performance of the federated averaging (FedAvg) and the federated proximity (FedProx) strategies on the heterogeneous data distributions scenario (Figure \ref{fig:ParametricStudy_Overview}-B). For the FedProx strategy, we vary the value of the regularization parameter $\mu$ from $1e-5$ (weak regularization) to $1e-2$ (strong regularization), as described in Figure \ref{fig:ParametricStudy_Overview}-D.

\subsubsection{Differential Privacy}
We evaluate the two differential privacy mechanisms described in Sec. \ref{sec:mm:differential_privacy} using two sites with a homogeneous data distribution of 50/50\% samples per site.
In the case of the Differential Privacy-Stochastic Gradient Descent (DP-SGD), we systematically varied the noise multiplier and gradient norm parameters according to Figure \ref{fig:ParametricStudy_Overview}-E. 
For the Differential Privacy Sparse Vector Technique (DP-SVT) algorithm, we varied the gradient query noise ($\epsilon_1$), the gradient clipping and the portion to release (Q: percentage of model parameters shared at each communication round) as presented in Fig. \ref{fig:ParametricStudy_Overview}-F.

%% file: sections/results.tex
\section{Results and Discussion}

\subsection{Communication rounds}

We assess the impact of the number of communication rounds and local training epochs by training federated models up to 64 accumulated training epochs using different rounds vs. epochs configurations. The parameters changing in this experiment are: number of samples per site ( \ref{fig:ParametricStudy_Overview}-A), number of samples with positive labels (Figure \ref{fig:ParametricStudy_Overview}-B) and communication frequencies (Figure \ref{fig:ParametricStudy_Overview}-C). All models are trained using the FedAvg federation strategy with no differential privacy (DP). We assessed the resulting model performance on a hold-out test set (3'236 samples) using the AUROC and AUPRC metrics. The goal of the experiment is to understand the trade-offs between communication rounds and local training epochs and to understand how to optimize these parameters for different data distributions.

\begin{figure}[h]
    \centering
    \begin{subfigure}[b]{0.49\textwidth}
        \includegraphics[width=\textwidth]{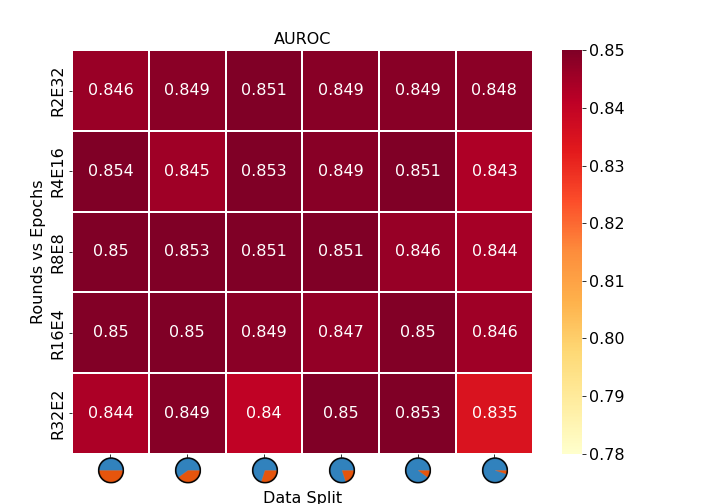}
        \label{fig:homogeneous_auroc}
    \end{subfigure}
    \hfill
    \begin{subfigure}[b]{0.49\textwidth}
        \includegraphics[width=\textwidth]{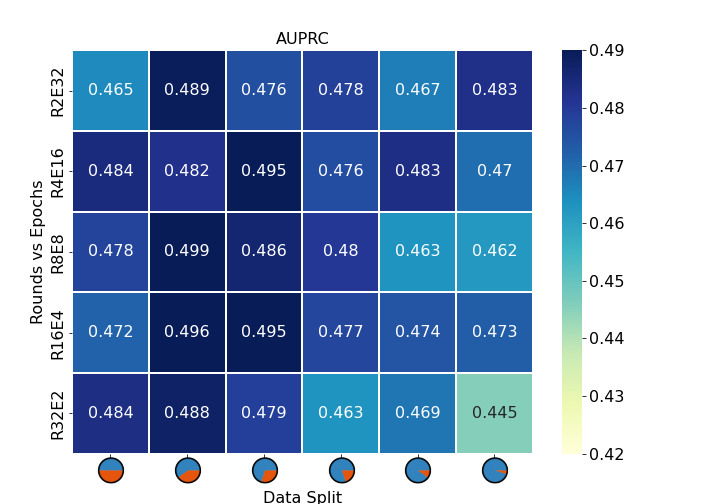}
        \label{fig:homogeneous_auprc}
    \end{subfigure}
    \begin{subfigure}[b]{0.49\textwidth}
        \includegraphics[width=\textwidth]{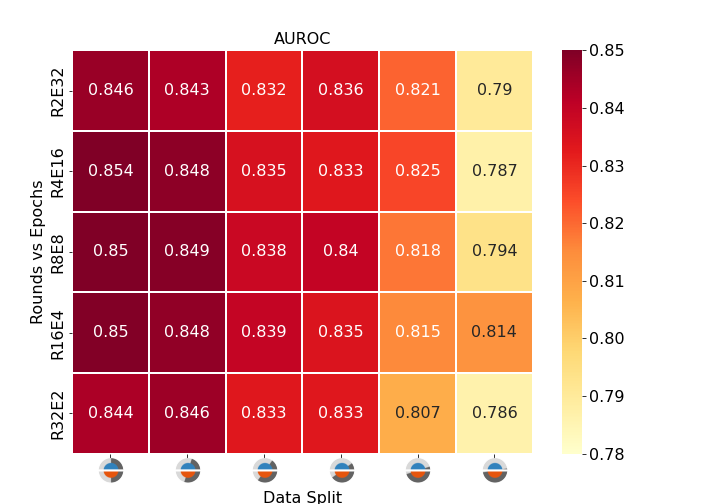}
        \label{fig:heterogeneous_auroc}
    \end{subfigure}
    \hfill
    \begin{subfigure}[b]{0.49\textwidth}
        \includegraphics[width=\textwidth]{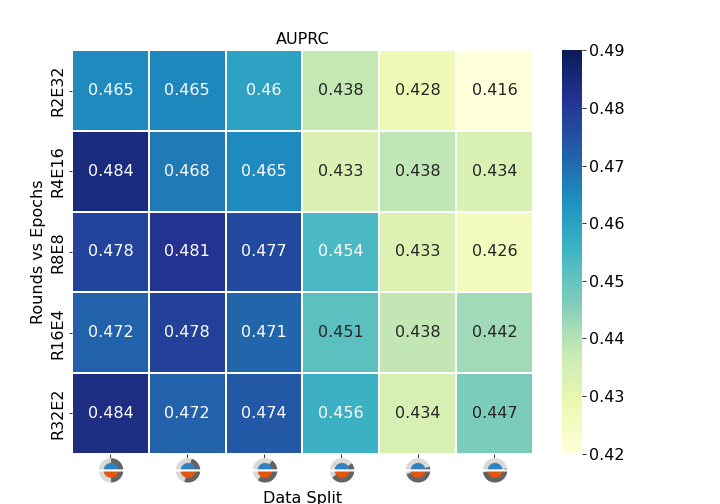}
        \label{fig:heterogeneous_auprc}
    \end{subfigure}
    \caption{Federated models performance - Communication rounds vs local training epochs: the models were trained in the following communication round/epoch configurations up to an accumulated 64 epochs: 32/2, 4/16, 8/8, 4/16, 2/32 on all homogeneous and heterogeneous data distributions. We used federated averaging (FedAvg) as federation strategy. Top panels: AUROC and AUPRC-based model performance on homogeneous data distributions. Bottom panels: AUROC and AUPRC-based model performance on heterogeneous data distributions.}
    \label{fig:heatmaps}
\end{figure}

Figure \ref{fig:heatmaps} shows the AUROC (red heat maps) and the AUPRC (blue heat maps) performance for federated models trained using different data distributions (x-axis) and different communication rounds / local training epochs (y-axis). The upper row of heat maps shows the federated model performance for an increasingly unbalance number of samples per site, starting from a 50/50\% to a 5/95\% distribution of samples. The heat maps show that the federated averaging (FedAvg) aggregation strategy can achieve similar performances (both AUROC and AUPRC) independent of the distribution of samples among sites. 
The bottom row of Figure \ref{fig:ho_data_distribution} shows the same heat maps for the different heterogeneous data distributions, where the number of samples per site is kept constant, but the number of positive samples is modified, going from a 50/50\% to a 5/95\% split of the positive samples across the two sites. The heat maps suggest that federated averaging (FedAvg) aggregation strategy incurs performance degradation (both for AUROC and AUPRC) when the label distributions become more and more unbalanced between the two sites. Furthermore, Figure \ref{fig:ho_data_distribution} bottom right panel shows that the AUPRC performance degradation is also dependent on the number of communication rounds, where setups with higher number of communication rounds scored higher results than the ones with fewer communication rounds.

\subsection{Data Distributions}
\label{sec:data_distributions}
We performed all following federated training experiments using 8 communication rounds, and at each communication round the model was trained locally for 8 epochs (same as middle row of heat-maps in Fig. \ref{fig:heatmaps}). In general, federated models outperformed the baseline models (trained exclusively on the on-site data) on the hold-out test set, for most of the evaluated data distribution scenarios (Figure \ref{fig:ho_data_distribution}, \ref{fig:he_data_distribution}). 
In particular, with an increasingly unbalanced number of samples and positive label sample distributions, the performance of the baseline models significantly diverged from each other (between Site \# and Site \#2). Besides the federated averaging (FedAvg) aggregation strategy, we also assessed the federated proximity (FedProx) aggregation strategy with different regularization parameters $\mu$ for the heterogeneous data distribution case. 


\subsubsection{Homogeneous Data Distribution}
\label{sec:data_distributions:homogeneous}
\begin{figure}[h!]
    \centering
    \begin{subfigure}[b]{0.49\textwidth}
        \includegraphics[width=\textwidth]{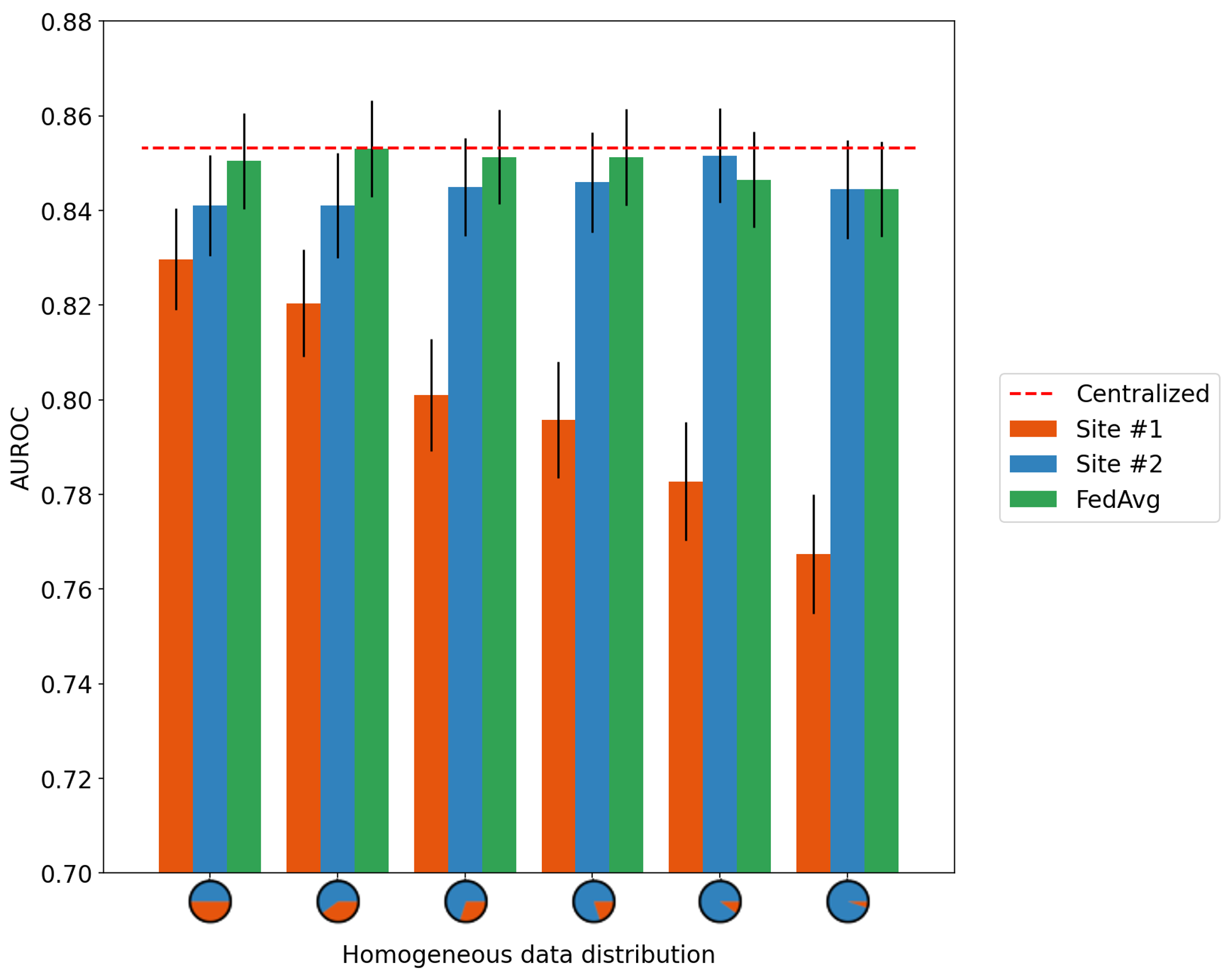}
    \end{subfigure}
    \hfill
    \begin{subfigure}[b]{0.49\textwidth}
        \includegraphics[width=\textwidth]{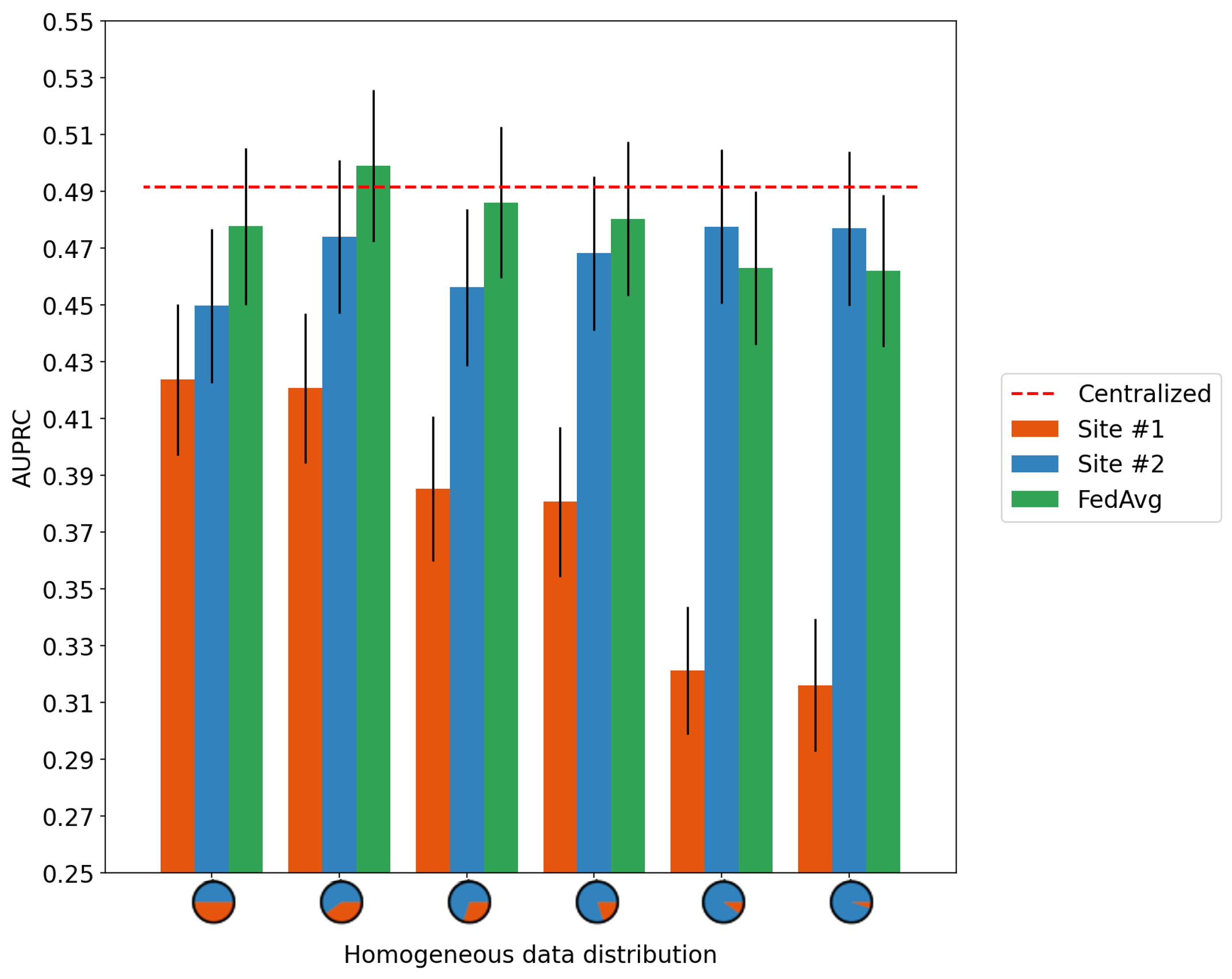}
    \end{subfigure}
    \caption{Federated vs. silo-based training on homogeneous data splits - model performances (AUROC and AUPRC) on a hold-out test set.  Local per-site training (red and blue columns) was performed on the individual sites, while the federated training (green columns) was performed using the simulated sites described in Figure \ref{fig:ParametricStudy_Overview}. The red-dashed line displays the performance of the model trained in the optimal case scenario (all data available for training in a silo-based manner).}
    \label{fig:ho_data_distribution}
\end{figure}

In this set of experiments, we distributed the available patients between Site \#1 and Site \#2 according to data distribution described in Fig. \ref{fig:ParametricStudy_Overview}-A, while keeping the ratio of positive to negative labels on each site constant. We used the FedAvg strategy to train the federated model. Figure \ref{fig:ho_data_distribution} shows that the federated models outperform the locally trained baseline in almost all cases, both for the AUROC, as well as the AUPRC metric. However, Figure \ref{fig:ho_data_distribution} also shows, that the best performance improvement for federated models is achieved when sites have a similar number of samples and positive labels.

\subsubsection{Heterogeneous Data Distribution}
\label{sec:data_distributions:heterogeneous}
We simulated differences in label distribution between the two sites as follows: given N=100 patients with positive labels in the total dataset, we progressively shifted patients with positive labels from one site to the other. Starting with a 50:50\% split between Site \#1 and Site \#2, we moved positive patients from Site 1 to Site 2, ending with a 5:95\% split. Differences in patient counts were made up by moving negatively labeled patients from Site \#2 to Site \#1. 

\begin{figure}[h!]
    \centering
    \begin{subfigure}[b]{0.49\textwidth}
        \includegraphics[width=\textwidth]{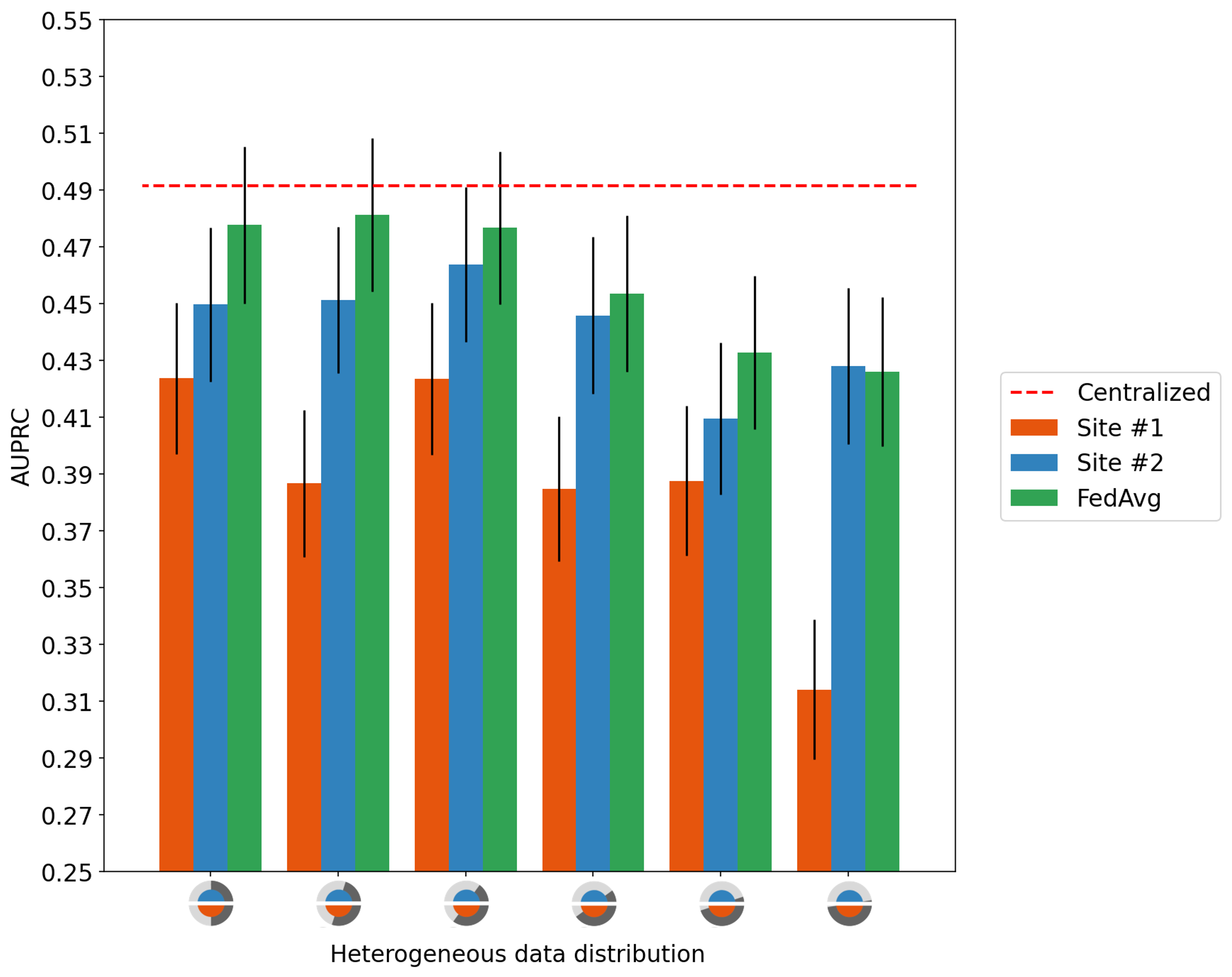}
    \end{subfigure}
    \hfill
    \begin{subfigure}[b]{0.49\textwidth}
        \includegraphics[width=\textwidth]{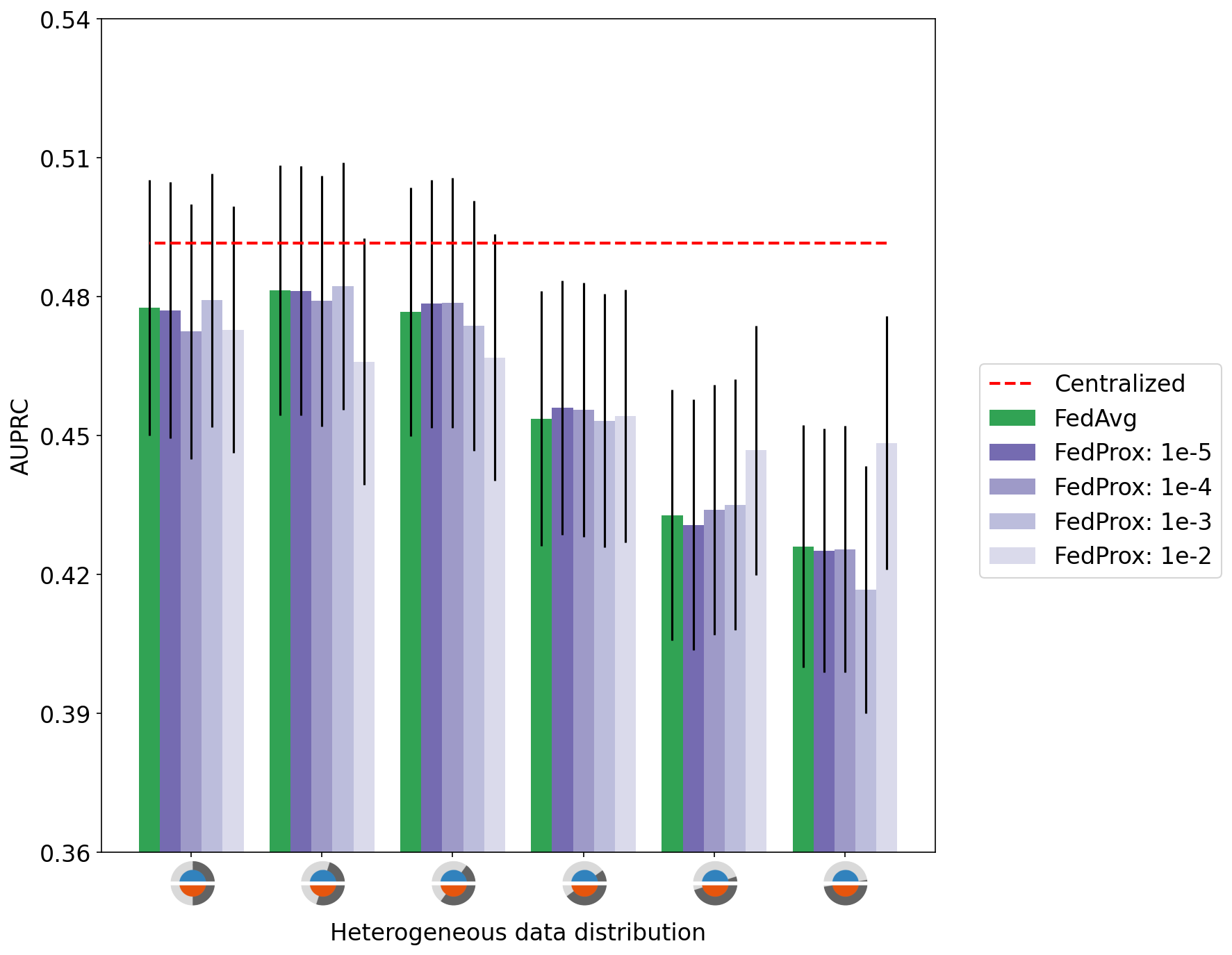}
    \end{subfigure}
    
    \caption{Silo, Federated averaging (FedAvg) and Federated proximity (FedProx) training on heterogeneous data splits - model performance (AUPRC) on a hold-out test set. Left panel: Silo training (red and blue columns) was performed on the individual sites, while the federated training using FedAvg (green columns) was performed using the simulated sites described in Figure \ref{fig:ParametricStudy_Overview}. Right panel: Federated Averaging (green columns) and federated proximity (violet columns with different transparency based on the regularization parameter $\mu$). The explored data splits and choice of the $\mu$ parameter are defined in Figure \ref{fig:ParametricStudy_Overview}-B,D. The performance of the model trained in the optimal case scenario (all data available for training in a silo-based manner) is displayed with a red-dashed line.}
    \label{fig:he_data_distribution}
\end{figure}

The federated model was trained with both the Federated Averaging (FedAvg) and the Federated Proximity (FedProx) algorithms, with the latter being more beneficial for data scenarios with differences in label ratios. In the case of FedProx, we varied the parameter $\mu$ (regularization parameter) according to Fig. \ref{fig:ParametricStudy_Overview}-D. In this set of experiments, we used only the AUPRC metric to compare the different scenarios due to the change in label distribution between the different data distributions. The goal of this experiment is to compare the performance of two federation strategies, in particular what performance improvements can be generated by using the FedProx algorithm.
Figure \ref{fig:he_data_distribution} shows, that for almost all heterogeneous data splits explored, federated proximity could further improve the performance of the federated averaging-based model. However, Figure \ref{fig:he_data_distribution} right shows that in order to achieve this increase in performance, the model developer has to conduct a hyper-parameter search for the optimal federated proximity parameter $\mu$.

\begin{figure}
    \centering
    \includegraphics[width=\textwidth]{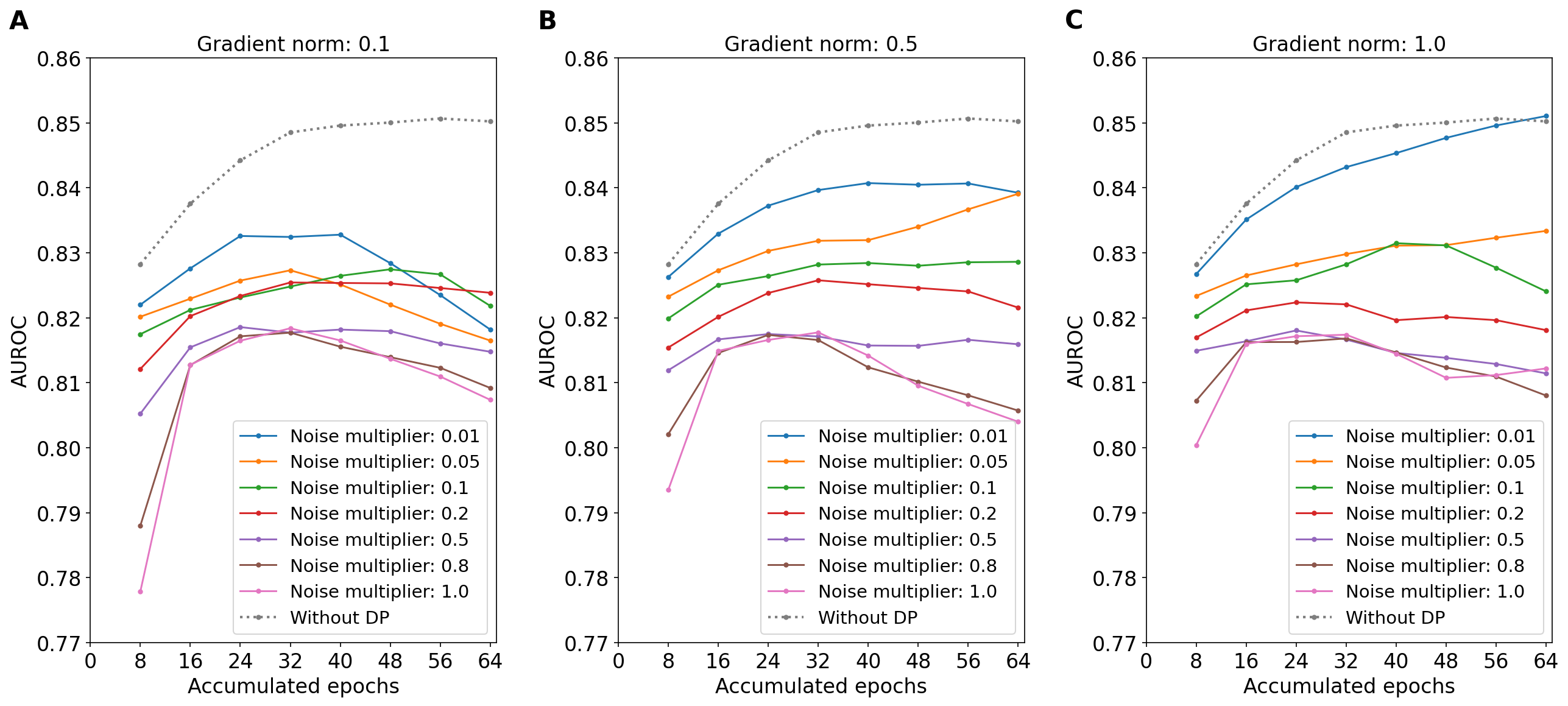}
    
    \caption{Differential Privacy – Stochastic Gradient Descent: AUROC performance of the federated model on a hold-out test set for three different gradient norms (panels A-C) and different noise multipliers (individual lines within each panel).}
    \label{fig:dp-sgd}
\end{figure}

\subsection{Differential Privacy}
\label{sec:results:dp}
As part of the parametric study, we also evaluated two different differential privacy (DP) techniques. While the DP-Stochastic Gradient Descent (DP-SGD) mechanism directly perturbs the model weights during each back-propagation step, the DP-Sparse Vector Technique (DP-SVT) only applies perturbation to the trained model weight prior to communicating them to the server using selective sharing and the sparse vector technique (SVT). 
The key motivation for our choice of these techniques originates from the problems that each technique solves, the former approach allows for a lower privacy cost (due to a tighter estimation of its privacy cost upper bound), while the latter provides flexibility to developers as it can be used with any type of model (beyond neural networks) and can be applied post-training (beneficial for open platforms, where developers can write their own models and it's difficult to enforce the use of DP-SGD). 

\subsubsection{Stochastic Gradient Descent (DP-SGD)}
\label{sec:results:dp:sgd}
We evaluated the stochastic gradient descent (SGD)-based differential privacy mechanism by using a homogeneous data distribution with 50:50\% data allocation at each site (Fig. \ref{fig:ParametricStudy_Overview}-A, 50-50), as well as a training procedure consisting of 8 local training epochs and 8 communication rounds (Fig. \ref{fig:ParametricStudy_Overview}-C, R8E8). Furthermore, we varied the maximal gradient norm (clipping parameter) and noise multiplier parameter (Fig. \ref{fig:ParametricStudy_Overview}-E) of the differential privacy mechanism. The goal of this experiment is to understand how the gradient norm and noise multiplier interact and whether optimal combinations of the parameters can be achieved.

Figure \ref{fig:dp-sgd} shows that independent of the gradient norm used, larger noise multipliers result in lower model performance and slower model convergence. The different gradient norms tested (Figure \ref{fig:dp-sgd} A-C) influence the final models performance; when a large gradient norm is used (e.g., $1.0$) the model can achieve performance levels similar to federated models trained without differential privacy (for low noise multiplier levels).  

\subsubsection{Sparse Vector Technique (DP-SVT)}
\label{sec:results:dp:svt}
We evaluated the sparse vector technique (SVT)-based differential privacy mechanism by using a homogeneous data distribution with 50:50\% data allocation at each site (Fig. \ref{fig:ParametricStudy_Overview}-A, 50-50), as well as a training procedure consisting of 8 local training epochs and 8 communication rounds (Fig. \ref{fig:ParametricStudy_Overview}-C, R8E8). Furthermore, we varied the $\varepsilon_1$ noise parameter, gradient clipping, and the percentage of model parameters shared between the local and the federated site at each communication round (Q) according to Fig. \ref{fig:ParametricStudy_Overview}-F) of the DP-SVT algorithm.

\begin{figure}[h!]
    \centering
    \includegraphics[width=\textwidth]{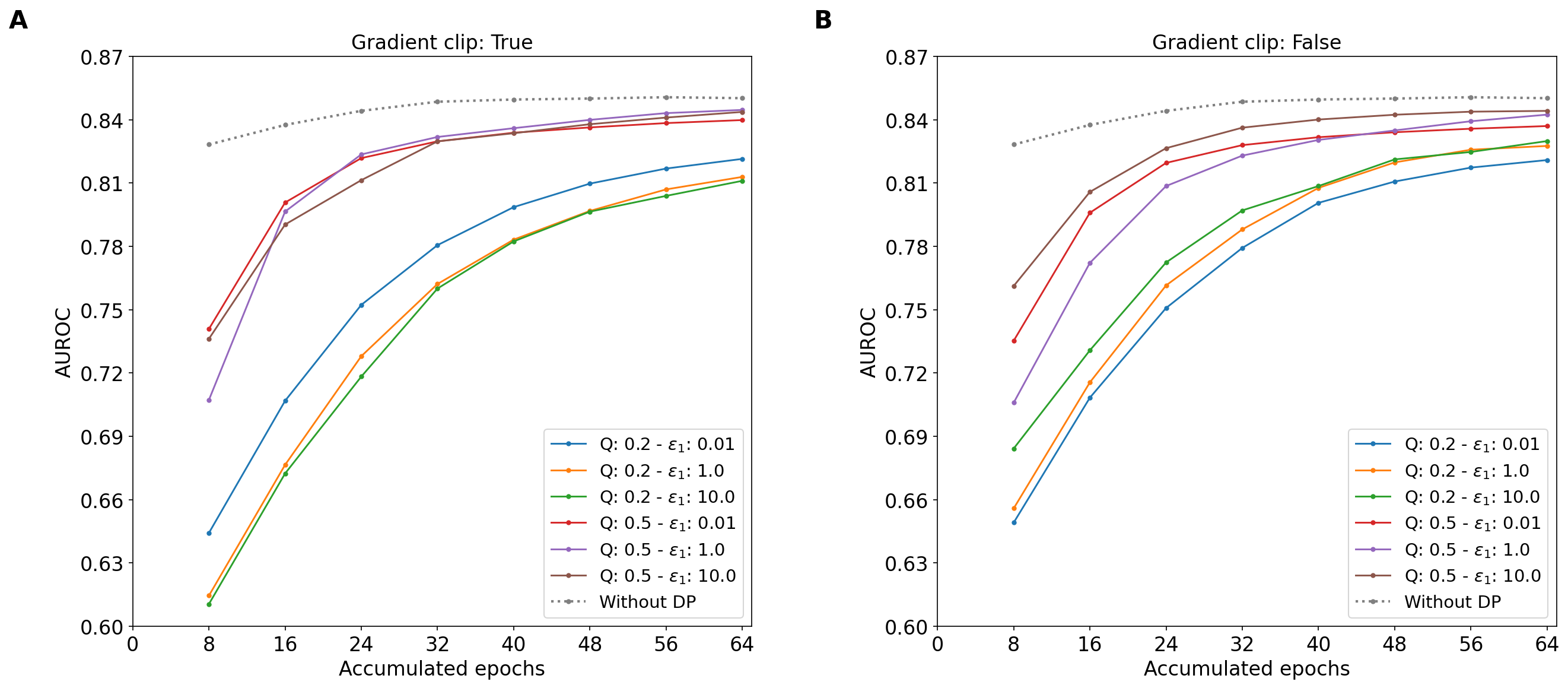}
    \caption{Differential Privacy – Sparse Vector Technique: Performance of the federated model on a hold-out test set. Panel A shows the performance convergence curve for different percentages of model parameters shared at each communication round (Q) and different noise parameters ($\varepsilon_1$) with gradient clipping enabled, while panel B shows the same set of parameters with gradient clipping disabled.}
    \label{fig:dp-svt}
\end{figure}

Figure \ref{fig:dp-svt} shows that sharing larger number of model parameters results in higher performance (red, violet and brown curves). Interestingly, increases in noise parameter (e1) did not result in lower model performance across the different parameter settings. Gradient clipping resulted in slightly lower model performance.

\subsection{Comparison}
We compared the performance (AUROC) and privacy leakage (measured by the parameter $\varepsilon$) for the two presented differential privacy mechanisms (DP-SGD and DP-SVT), by using a homogeneous data distribution with 50:50\% data allocation at each site (Fig. \ref{fig:ParametricStudy_Overview}-A, 50-50), as well as a training procedure consisting of 8 local training epochs and 8 communication rounds (Fig. \ref{fig:ParametricStudy_Overview}-C, R8E8). We used the federated averaging (FedAvg) strategy to train the federated model.
Furthermore, for each differential privacy mechanisms we varied its mechanism specific parameters according to Fig. \ref{fig:ParametricStudy_Overview}-E for DP-SGD and Fig. \ref{fig:ParametricStudy_Overview}-F for DP-SVT.

\begin{figure}[h!]
    \centering
    \begin{subfigure}[b]{0.49\textwidth}
        \includegraphics[width=\textwidth]{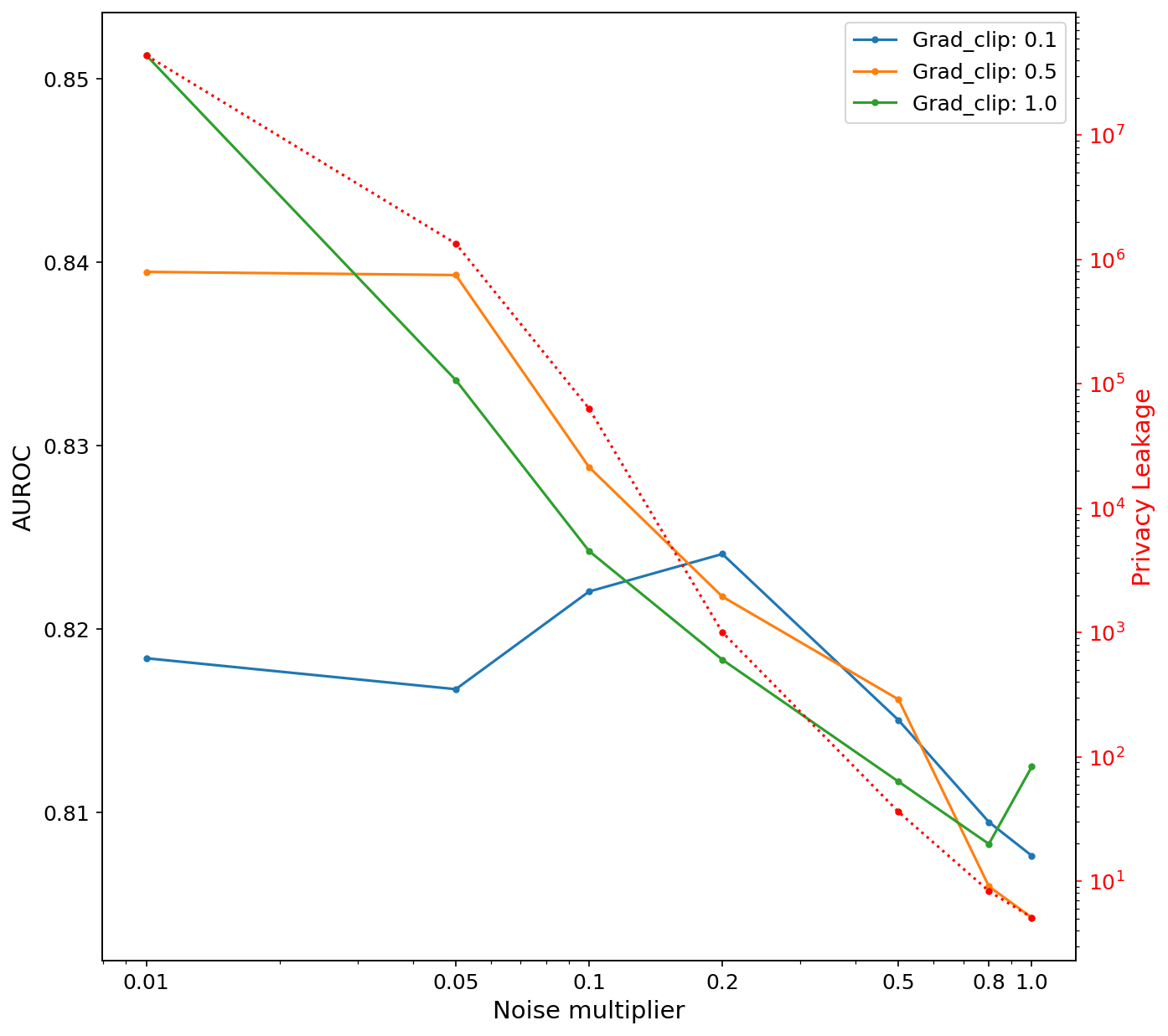}
        \label{fig:dp_sgd_noise_vs_auroc}
    \end{subfigure}
    \hfill
    \begin{subfigure}[b]{0.49\textwidth}
        \includegraphics[width=\textwidth]{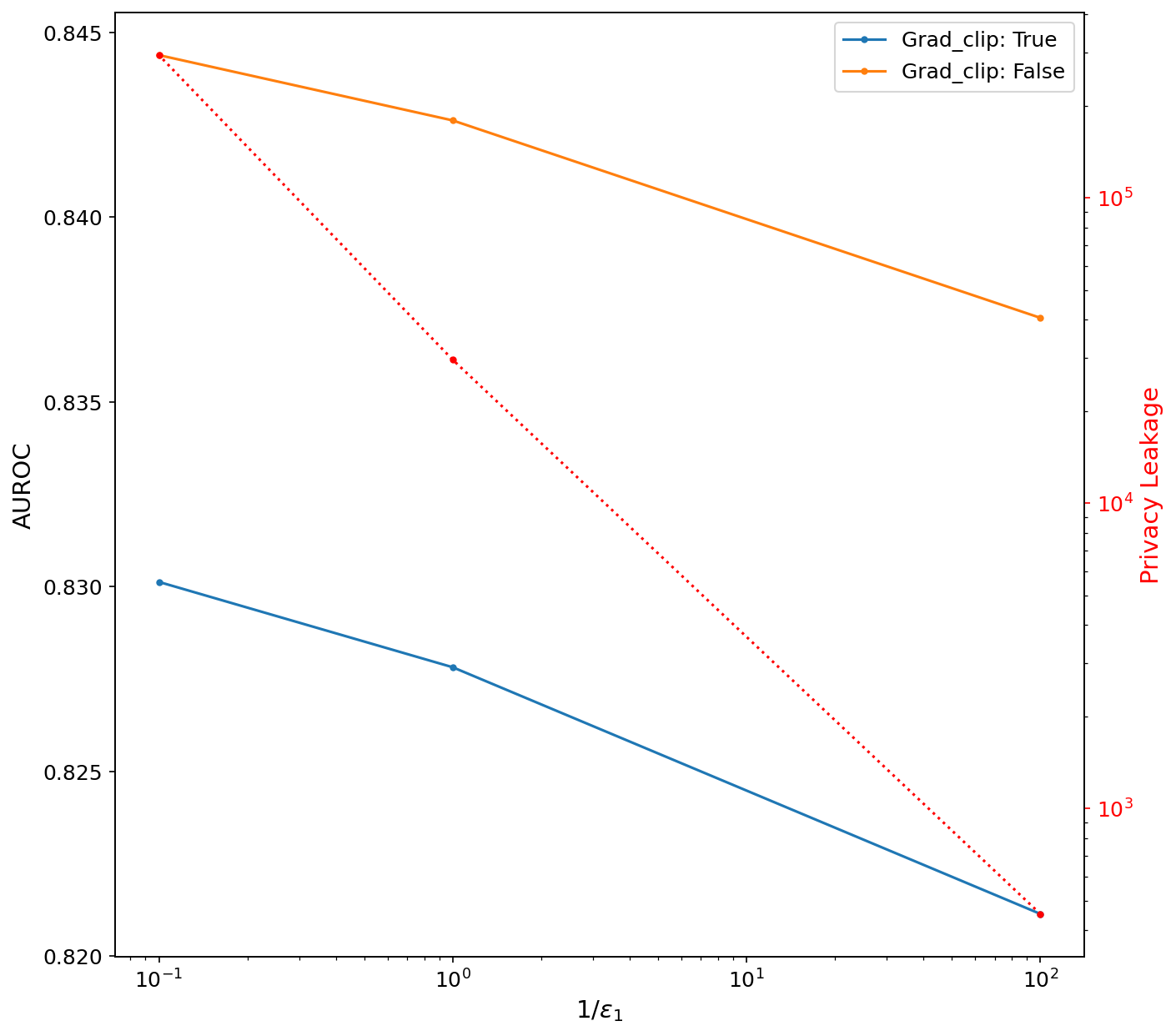}
        \label{fig:dp_svt_noise_vs_auroc}
    \end{subfigure}
    
    \caption{Differential Privacy - Performance vs. privacy leakage trade-off. Left panel: various noise multipliers and gradient clipping parameters for DP-SGD. Right panel: various $1/\varepsilon_1$ and gradient clipping parameters for DP-SVT.}
    \label{fig:dp_noise_vs_auroc}
\end{figure}

Figure \ref{fig:dp_noise_vs_auroc} shows the trade-off in performance when choosing larger noise multipliers (DP-SGD) and higher $1/\varepsilon_1$ (DP-SVT). Particularly of interest is the upper bound estimation of the privacy leakage (right y-axis), which shows that the privacy leakage and the noise multiplier have a roughly inverse linear relationship (in log-log scaled diagram). Therefore, choosing a larger noise multiplier can lead to a small AUROC performance decrease, while at the same time reducing the privacy leakage by orders of magnitude (Figure \ref{fig:dp_noise_vs_auroc} - Panel left). The same can be observed for the DP-SVT algorithm and its corresponding $\varepsilon_1$ parameter (Figure \ref{fig:dp_noise_vs_auroc} - Panel right). The AUROC performance decreases slightly when decreasing $\varepsilon_1$, while at the same time the privacy leakage decreases by two orders of magnitude. 

%% file: sections/conclusions.tex
\section{Discussion \& Conclusion}

We presented a trade-off analysis of the different parameters that influence the final performance of models trained on the open-source MIMIC-III dataset using federated learning algorithms, and differential privacy techniques.

In Section \ref{sec:data_distributions:homogeneous}, we presented the performance of federated models trained using two separate sites. Our results (Fig. \ref{fig:ho_data_distribution}) demonstrate a significant performance improvement in AUROC and AUPRC metrics for models trained using federated averaging (FedAvg) compared to models trained on local data (i.e. in silos training). Overall, the performance of the federated models is on average slightly below the performance of a model trained on the full dataset in one central location. The presented results are in line with the ones reported by Budrionis et al. \cite{Budrionis2021} on their performance analysis of the PySyft federated learning framework on the MIMIC-III dataset. They used 32 separate sites with same number of samples per site, and achieved a federated model performance close to a model trained on the full dataset, as shown in our experiments as well. 

In Section \ref{sec:data_distributions:heterogeneous} we showed the performance gains that can be achieved using FedProx compared to FedAvg when the data is distributed heterogeneously across the federated sites / nodes. These results are in line with the work by Li et al. \cite{Li2018}, which compared the training loss of FedAvg with FedProx and showed that the proximity term improves performance by stabilizing the training loss over different communication rounds. This is further highlighted in our work in Fig. \ref{fig:he_data_distribution}, where FedProx showed to be particularly useful for cases where the data distribution across sites is highly heterogeneous. As the data distribution approaches homogeneity, FedProx performs comparable to FedAvg.
Additionally, the choice of the proximity penalization parameter $\mu$ plays an important role in the performance of the final federated model. Therefore, further research on per-site optimization of the proximity parameter $\mu$ could lead to further performance improvement. Currently, a simple, yet computationally expensive strategy is to use grid-search (as we have done in our experiments) to optimize this hyper-parameter.

Furthermore, our analysis of the training epochs versus communication rounds in Fig. \ref{fig:heatmaps}, highlights the need for a strategy to select these two hyper-parameters before the model training starts because of their influence on the final models performance. A conservative choice of the parameters can be set by 
choosing a high number of communication rounds and a relatively small number of local training epochs, as this would avoid divergence of the model parameters at the different sites during the local training procedure. 

Lastly, we compared the performance of federated models trained with two separate differential privacy techniques aimed at reducing the privacy leakage through model memorization or adversarial attacks (in Section \ref{sec:results:dp}). The results suggest that using differential privacy would lead to a reduction in final models performance. However, we observe a higher noise injection leads to a relatively small reduction in model performance of ca. 5\%, while reducing the privacy leakage by up to 6 orders of magnitude at the same time (Fig. \ref{fig:dp_noise_vs_auroc}). 

Further work is required for quantifying the privacy risk for a single patient record when using differential privacy strategies. As of writing, the authors are not aware of legislation addressing the privacy risks deriving from sharing models trained on private data.

%% file: sections/backmatter.tex
\subsection*{Competing interests}
The authors declare that they have no competing interests.

\subsection*{Acknowledgements}
This work was funded by the Swiss Innovation Agency (Innosuisse) through the grant Nr. 42089.1.

\subsection*{Author's contributions}
ANH and MB contributed equally to the writing of the manuscript and to the development of the code base. FN contributed to the development of the machine learning model and training infrastructure. AA and MK provided scientific guidance, edited and reviewed the manuscript.

\section*{Acronyms}

\begin{tabular}{ l l }
 \textbf{IID} & Independent and identically distributed \\
 \\
 \textbf{FL} & Federated Learning \\
 \textbf{DP} & Differential Privacy \\
 \textbf{DP-SGD} & Stochastic Gradient Descent Differential Privacy \\
 \textbf{DP-SVT} & Sparse Vector Technique Differential Privacy \\
 \textbf{SVT} & Sparse Vector Technique \\
 \\
 \textbf{SGD} & Stochastic Gradient Descent \\
 \textbf{ReLU} & Rectified Linear Activation Unit \\
 \textbf{SVRG} & Stochastic Variance Reduced Gradient \\
 \textbf{FSVRG} & Federated Stochastic Variance Reduced Gradient \\
 \textbf{LSTM} & Long-Short Term Memory \\
 \\
 \textbf{ROC} & Receiver Operating Characteristic \\
 \textbf{PR} & Precision-Recall \\
 \textbf{AUC-ROC} & Area Under the Receiver Operating Characteristics Curve \\
 \textbf{AUROC} & Equivalent to AUC-ROC \\
 \textbf{AUC-PR} & Area under the Precision Recall Curve \\
 \textbf{AUPRC} & Equivalent to AUC-PR \\
 \textbf{RMSE} & Root mean-square error \\
 \textbf{MAE} & Mean absolute error \\
 \textbf{NLL} & Negative Log-Likelihood \\
 \\
 \textbf{ICD-10} & International Classification of Disease (version 10) \\
 \textbf{MIMIC-III} & Medical Information Mart for Intensive Care III \\
\end{tabular}

%% file: appendix/parametric_study.tex
\section{Parametric Study}
\label{sec:parametric_study_appendix}

\subsection{Federated proximity}

\begin{figure}[h!]
    \centering
    \includegraphics[width=0.5\textwidth]{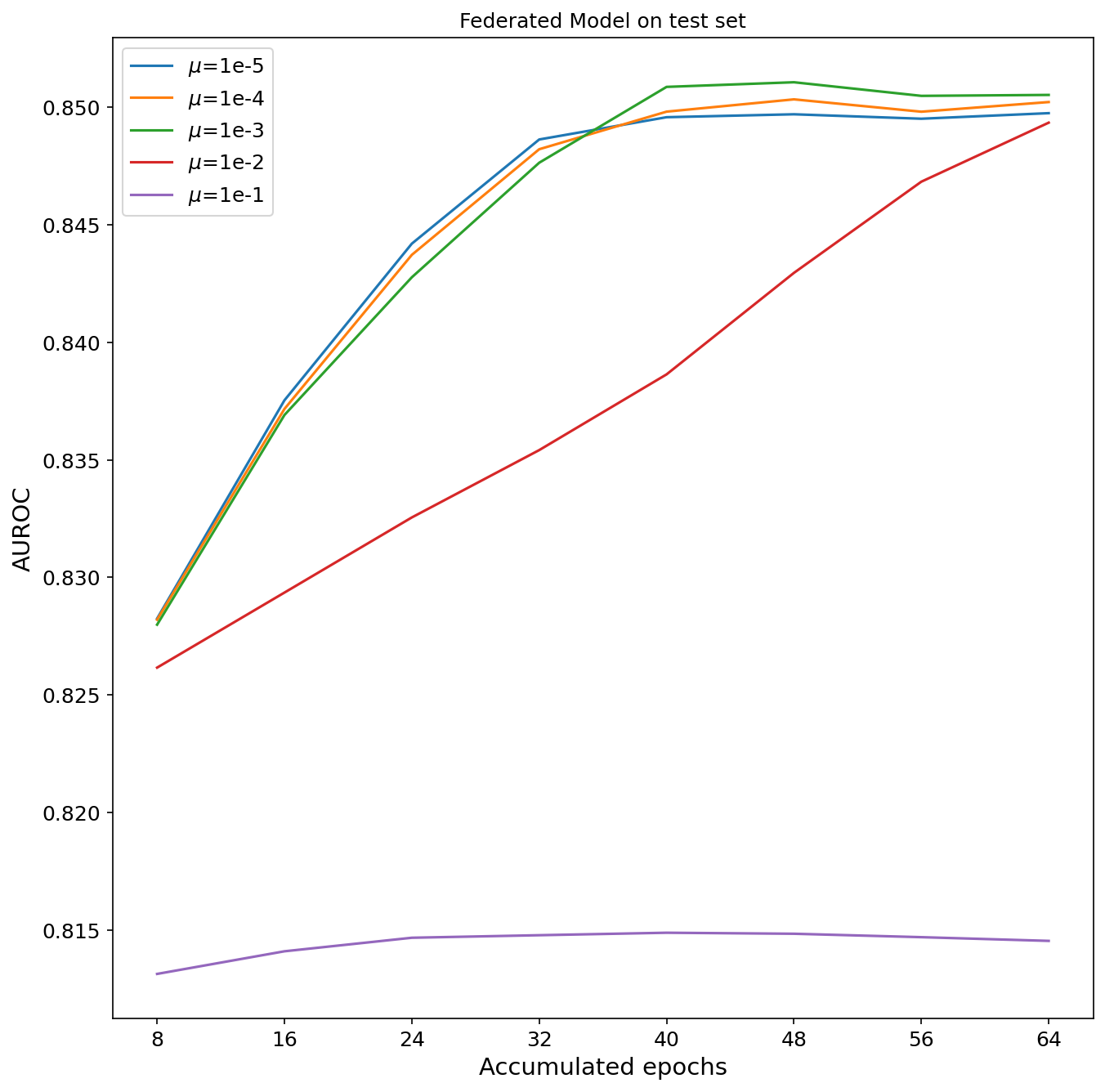}
    \caption{The convergence behavior of the federated learning model is strongly influenced by the choice of the parameter $\mu$. Large values hinder the learning procedure by over-enforcing proximity to the federated weights, while small values have a diminishing impact on the federated model convergence.}
    \label{fig:he_convergence}
\end{figure}

Figure \ref{fig:he_convergence} shows that larger values of the proximity parameter $\mu$ (larger than 1e-3) have a strong influence on the performance converge of the federated model. In particular, for a choice of $\mu=$1e-1, the proximity term in the loss calculation is hindering the training procedure, as we see that the performance is not improving and remains at its initial level. 

\begin{figure}
    \centering
    \includegraphics[width=\textwidth]{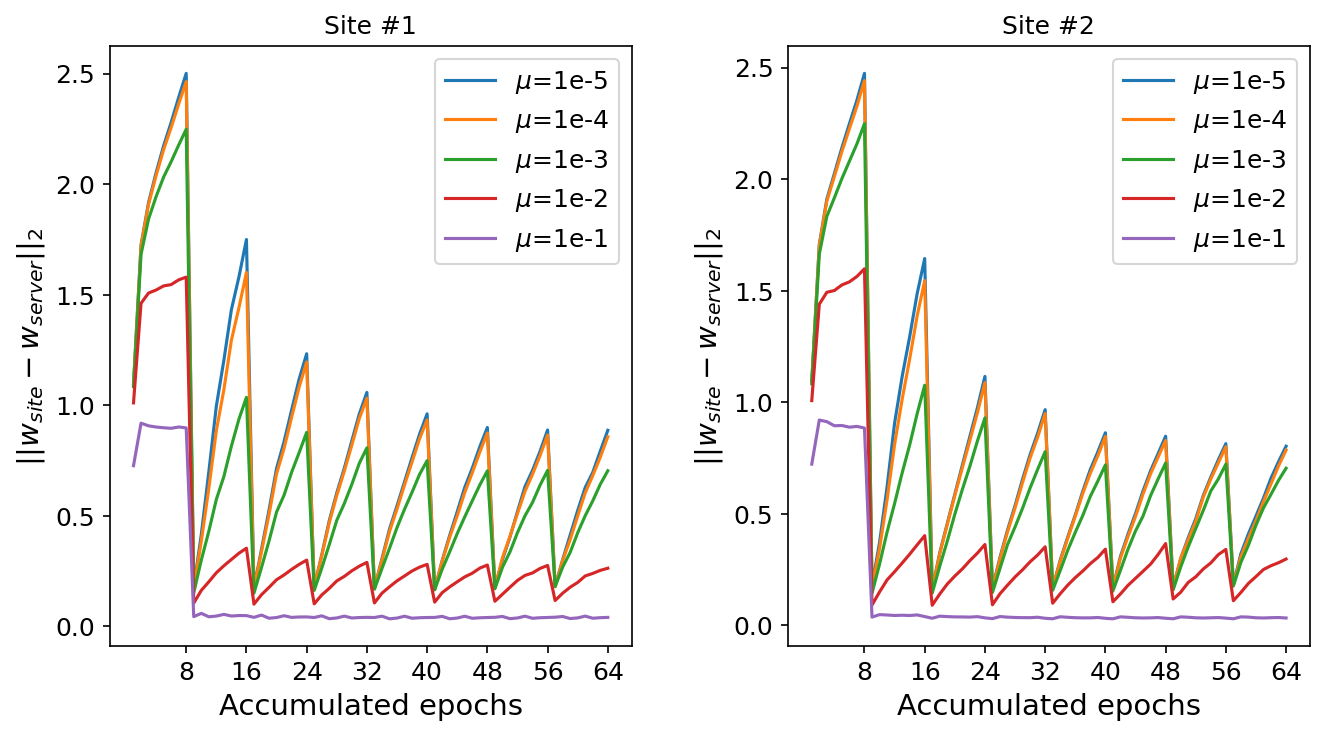}
    \caption{The figure displays the L2-norm component of the federated proximity loss term. The training was performed with 8 communication rounds and 8 local training epochs using sites with 50/50\% label distribution. The characteristic saw-tooth pattern is expected, as at each communication round the sites receive the new federated model weights and restart the training procedure}
    \label{fig:he_L2_norm}
\end{figure}

Figure \ref{fig:he_L2_norm} confirms this observation, by showing the magnitude of the local model weight changes in-between communication rounds. With increasing values of $\mu$ the magnitude of the weight changes is impacted, up to the point where it converges towards $0$ for federated models trained using a proximity parameters $\mu=$1e-1. The saw-tooth pattern visible in Figure \ref{fig:he_L2_norm} is expected, as the local model weights are expected to diverge from the federated model weights during local training. Upon receiving the new federated model weights, the magnitude of the difference resets to a lower level. 

Figure \ref{fig:he_convergence}, \ref{fig:he_L2_norm} reinforce the recommendation that a hyper-parameter search for the parameter $\mu$ is required during federated training to achieve optimal results. 

